\documentclass[journal]{IEEEtran}
\usepackage{amsmath,graphicx,cite,epsfig}
\usepackage{multirow}
\usepackage{booktabs}
\usepackage{float}
\usepackage{xspace}
\usepackage{amssymb}
\usepackage{amsmath}
\usepackage{amsthm}
\usepackage{subfigure}
\usepackage{caption}
\usepackage{threeparttable}
\usepackage{diagbox}
\usepackage{url}
\usepackage{bm}
\usepackage{bm}
\usepackage{amsfonts}
\newcommand{\ieno}{\textit{i.e.}}
\newcommand{\egno}{\textit{e.g.}}

\newcommand{\etal}{\textit{et al.}}
\usepackage[table,xcdraw]{xcolor}
\usepackage{algorithm}  
\usepackage{algorithmicx}  
\usepackage{algpseudocode}
\usepackage{amsmath}
\usepackage{bbding}
\ifCLASSINFOpdf

\else

\fi

\begin{document}
%\title{FreAM: Frequency Alignment and Movement based Unsupervised Domain Adaption for Non-reference Image Quality Assessment}
%
%\title{FreqAlign: Excavating Perception-oriented Transferability via Frequency Alignment}
%\title{FreqAlign: Frequency Alignment }
%\title{FreqAlign: Frequency Alignment for the Unsupervised Domain Adaptation of Blind Image Quality Assessment}
%\title{Perception-oriented Domain Adaptation}
\title{A Close Look at Few-shot Real Image Super-resolution from the Distortion Relation Perspective}


\author{Xin Li,~\IEEEmembership{Student Member, IEEE}, 
Xin Jin, Jun Fu, Xiaoyuan Yu, Bei Tong, Zhibo Chen,~\IEEEmembership{Senior~Member,~IEEE}% <-this % stops a space
\thanks{This work was supported in part by NSFC under Grant U1908209, 61632001. (Corresponding authors: Zhibo Chen and Xin Jin.)}
\thanks{X. Li, J. Fu, and Z. Chen are with the CAS Key Laboratory of Technology in Geo-Spatial Information Processing and Application System, University of Science and Technology of China, Hefei 230027, China (e-mail: lixin666@mail.ustc.edu.cn; fujun@mail.ustc.edu.cn; chenzhibo@ustc.edu.cn). X. Jin is with the Eastern Institute for Advanced Study (e-mail: jinxin@eias.ac.cn). X. Yu and B. Tong are with Huawei Cloud (e-mail: yuxiaoyuan@huawei.com, tongbei@huawei.com).
}}
\maketitle
\definecolor{mygray}{gray}{.9}
\begin{abstract}
Collecting amounts of distorted/clean image pairs in the real world is non-trivial, which seriously limits the practical applications of these supervised learning-based methods on real-world image super-resolution (RealSR). Previous works usually address this problem by leveraging unsupervised learning-based technologies to alleviate the dependency on paired training samples. However, these methods typically suffer from unsatisfactory texture synthesis due to the lack of supervision of clean images. To overcome this problem, we are the first to have a close look at the under-explored direction for RealSR, \ieno, few-shot real-world image super-resolution, which aims to tackle the challenging RealSR problem with few-shot distorted/clean image pairs. Under this brand-new scenario, we propose Distortion Relation guided Transfer Learning (DRTL) for the few-shot RealSR by transferring the rich restoration knowledge from auxiliary distortions (\ieno, synthetic distortions) to the target RealSR under the guidance of distortion relation. 
Concretely, DRTL builds a knowledge graph to capture the distortion relation between auxiliary distortions and target distortion (\ieno, real distortions in RealSR). Based on the distortion relation, DRTL adopts a gradient reweighting strategy 
to guide the knowledge transfer process between auxiliary distortions and target distortions. 
In this way, DRTL could quickly learn the most relevant knowledge from the synthetic distortions for the target distortion.
We instantiate DRTL with two commonly-used transfer learning paradigms, including pre-training and meta-learning pipelines, to realize a distortion relation-aware Few-shot RealSR. Extensive experiments on multiple benchmarks and thorough ablation studies demonstrate the effectiveness of our DRTL.
\end{abstract}

\begin{figure}
	\centering
	\includegraphics[width=0.9\linewidth]{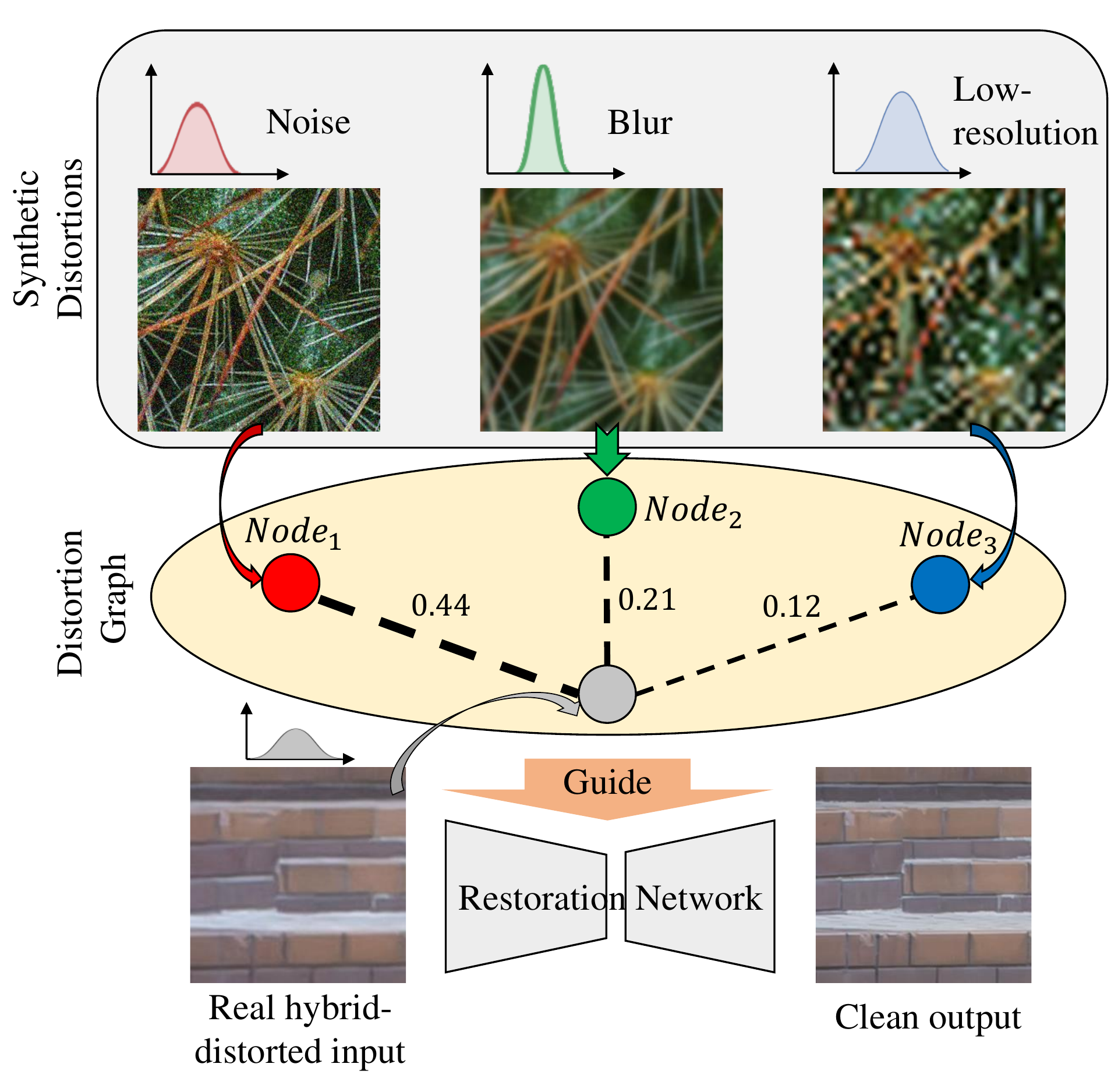}
	\caption{Illustration of the proposed Distortion Relation guided Transfer-Learning (DRTL). We introduce a distortion relation graph to capture the relationship (\ieno, 0.44, 0.21 and 0.12 shown above) between auxiliary distortions and real hybrid distortion (\ieno, RealSR), Then, we use these relations to adaptively guide the knowledge transfer process for real super-resolution.}
	\label{fig:motivation}
\end{figure}
% \begin{figure*}
%     \centering
%     \includegraphics[width=0.95\textwidth]{./fig1/sample2.png}
%     \caption{Examples of the domain shift between synthetic dataset and authentic dataset. The IQA metric trained on the synthetic Kadid10k dataset suffers from a severe performance drop when predicting the quality of images from the authetic KonIQ-10K dataset.}  
%     %ResNet-18 trained on $synthetic dataset Kadid10k   %}
%     %ResNet-18 trained on Kadid10k predict the quality of images in KonIQ-10k.}
%     \label{fig:teaser}
% \end{figure*}

% \fi
\IEEEpeerreviewmaketitle

\section{Introduction}
\label{sec:introduction}
%第一段UDA 和UDAIQA，
%第二段frequency UDA
%第三段freAM
 \IEEEPARstart{I} mage restoration (IR) task~\cite{tao2018scale,kupyn2019deblurgan,zhang2017beyond,moran2020noisier2noise,yu2016deep,xia2023diffIR,liang2021swinirIR,conde2023swin2srIR,wang2022uformerIR,deng2020detail,wang2019spatial,li2023learningCausalIR,li2020learningFDRNet,liu2020lira} aims to restore high-quality images from their degraded low-quality counterparts. The degradation process typically consists of a series of different distortions, such as low-resolution~\cite{dong2015image,dong2016accelerating}, motion blur \cite{tao2018scale,kupyn2019deblurgan}, noise \cite{zhang2017beyond,moran2020noisier2noise}, compression artifacts,  \cite{yu2016deep,li2020multi,li2023hst}, raining \cite{deng2020detail,wang2019spatial}, and etc. Although the fast-developed deep learning has significantly promoted the advancement of image restoration (IR) techniques, the success of them usually relies on training with a large-scale dataset. As a result, numerous image restoration datasets that contain various distortions have been proposed and deeply explored/studied, such as DIV2K \cite{agustsson2017ntire}, Flick2K~\cite{timofte2017ntireFlick2K}, Gopro \cite{nah2017deep}, and DID-MDN \cite{zhang2018density}. However, the degradations of these datasets are typically synthetic and are different from the real-world ones. To access real-world image restoration, the real-world image super-resolution (RealSR) task is proposed as the pioneer for real-world image restoration, where low-resolution (LR) images suffer from severe hybrid distortions, such as low-resolution, blur, noise, and compression artifacts.

 %As the typical task in real image 
 
 %Real-world image super-resolution~\cite{cai2019toward} (\ieno, RealSR) is the typical case of real image restoration, where low-resolution (LR) images suffer from severe hybrid distortions, such as blur, noise, and compression artifacts.      
%As the typical task of real-world image restoration, real image super-resolution (RSR) has attracted a large of attention in recent years. 

Although some learning-based super-resolution (SR) approaches~\cite{chen2020prePIT,dai2019second,zhang2018residual} that trained on the above-mentioned synthetic datasets have achieved a great success when handling the synthetic degradations (\ieno, bicubic degradation), those image SR systems hardly perform well on the real-world scenarios due to the large distribution gap between synthetic distortions and real-world distortions~\cite{wei2019semi,yasarla2020syn2real,li2021learning}. Furthermore, re-collecting large distorted/clean image training pairs from the real world is also non-trivial. 
The above-mentioned two disgusting weaknesses of the existing image SR algorithms seriously limit their practical applications and hurt their industrial values. To alleviate the dependency for clean distortion/clean image training pairs, some studies~\cite{ulyanov2018deepDIP,du2020learningInvarintIR,wang2021realESRGAN,fei2023generativeDiffIR,mou2022metricRealSR} introduce unsupervised learning to real image restoration. However, these methods typically suffer from unsatisfactory/spurious textures synthesis due to the lack of clean image supervision. Compared with a purely unsupervised solution, Few-shot RealSR is a potentially under-explored scheme, which only exploits few-shot distorted/clean image pairs and is  more feasible to tackle this challenging RealSR problem.

Previous works have explored different transfer learning strategies~\cite{chen2021metaMeta-baseline,rusu2018metaMLEO,finn2017modelMAML,snell2017prototypical,zhou2022conditionalCoop,lu2021simplerFew-shotSeg,lu2022predictionFew-shotSeg} for few-shot problems in the high-level tasks,  extracting helpful knowledge from auxiliary tasks to target task. Among them, Pre-training and Meta-learning are two representative technologies in few-shot learning, which are also preliminarily explored in image super-resolution tasks but with different purposes.
For instance, IPT \cite{chen2020prePIT} introduces a large-scale pre-training dataset to improve the restoration performance w.r.t the target distortion. Soh \textit{et al.}~\cite{soh2020meta} propose a meta-learning-based method to implement the fast adaptation for zero-shot super-resolution tasks, achieving a SOTA performance. However, these works still focus on handling the \textbf{synthetic} distortions and are not designed for few-shot learning of image restoration, which is not suitable for a more challenging Few-shot RealSR problem.

%In this paper, unlike the existing transfer learning-based IR methods that are typically designed for dealing with synthetic distortions, 
In this paper, we are the first to have a close look at  the challenging few-shot RealSR problem, where we aim to transfer the rich restoration knowledge from auxiliary distortions (\ieno, synthetic distortions) to the target distortion (\ieno, real-world super-resolution) with few-shot distorted/clean real-world image pairs. However, there are existed a large domain gap between the auxiliary distortions and the target distortion.
%the auxiliary tasks (\ieno, synthetic distortions) and target tasks (\ieno, real super-resolution) are different and exist a large domain gap. 
Therefore, we face the key issue required to be solved: ``How to capture the proper prior knowledge for target distortion from auxiliary distortions?''
Different from high-level classification tasks, of which the relationship between different tasks can be modeled with a simple clustering like KNN \cite{krishna1999genetic}. To capture the relation between different distortions adaptively, we propose to learn a distortion relation graph, which consists of two essential factors: Nodes (\ieno, distortion embeddings) and Edges (\ieno, the similarity between distortions) as shown in Figure \ref{fig:motivation}. We introduce a distortion relation network (DRN) to extract/learn the expected distortion relation embeddings. To make the DRN general for arbitrary real image super-resolution, we design a prior knowledge memory bank to store the learnable distortion relation priors from seen auxiliary tasks. Given an arbitrary real distorted sample, it can traverse the prior knowledge memory bank to acquire the required distortion embeddings as in GCN~\cite{kipf2016semi}.
%we set the prior knowledge memory bank to strage the 
%To generate nodes for each distortion, we design the distortion-ware graph network (DGN), which utilize the residuals between distorted and clean pairs as input of DGN. To establish the relationship between different distortions, we set the distortion relationship memory bank to store the priors of distortions.  Then we utilize graph convolution to further extract the relationship between different distortions.

After obtaining the required distortion embeddings, we compute the edges (that represent the distortion similarity) between different distortions with cosine similarity as the relation. This relation can measure the transferability of specific auxiliary distortion to the target distortion. With the guidance of distortion relation, we propose the novel \textbf{D}istortion \textbf{R}elation guided \textbf{T}ransfer \textbf{L}earning (DRTL), which utilizes the distortion relation to revise the optimization direction by the gradient reweighting strategy in the transfer learning process and let the learned knowledge from auxiliary distortions be more available for target distortion. 
% To utilize the distortion relation to guide the knowledge transfer process in Few-shot RealSR, 
%Since the different auxiliary distorted samples contain different degrees of useful priors for target real distortion, 
%we propose a Distortion-Relation guided Transfer Learning (DRTL). 
We instantiate our DRTL with two commonly-used transfer learning paradigms in few-shot learning, \ieno, pre-training, and meta-learning paradigms (dubbed as DRTL$_{p}$ and DRTL$_{m}$). 
%Specifically, we integrate the distortion similarity that is inferenced out from distortion-graph into the optimization loop of pre-training/MAML~\cite{finn2017model} with gradient weighting.
Extensive experiments on several typical backbones have demonstrated the effectiveness of our DRTL on few-shot real image super-resolution. A more thorough analysis of the distortion relation and few-shot RealSR is provided in the ablation studies.

% Also, we are the first to introduce a distortion graph to explore the relationship between synthetic distortions and real-world distortions.

%% 描述contributions
The main contributions of this paper can be summarized as follows:

\begin{itemize}
    \item To our knowledge, we are the first to have a close look at a brand-new RealSR task, \ieno, few-shot real image super-resolution (RealSR), which aims to transfer the rich knowledge from auxiliary distortions (\ieno, synthesized distortions) to the target RealSR task with few-shot real-world distorted/clean image pairs.

    \item We propose the distortion relation graph to measure the transferability between distortions, where a prior knowledge memory bank is exploited to store the learned knowledge relation priors from seen auxiliary distortions. 
   
    %\item  We propose a distortion relation graph to guide the few-shot real image super-resolution from the perspective of exploring the relationship between synthetic distortions and real-world distortion.
   
    \item With the guidance of distortion relation,  we propose the novel \textbf{D}istortion \textbf{R}elation guided \textbf{T}ransfer \textbf{L}earning (DRTL) for few-shot RealSR, which exploits the distortion relation to reweight optimization direction in the knowledge transfer process, and obtains more reliable and rich restoration knowledge for target distortion from auxiliary distortions with few-shot real distorted/clean image pairs.    
    % knowledge transfer/training strategy embedded with distortion-relation prior guidance for FS-RSR, \ieno, Distortion-Relation guided Transfer Learning (DRTL).  
    \item We instantiate our DRTL with two popular transfer learning paradigms, including pre-training and meta-learning. Extensive experiments on multiple typical restoration networks and these two transfer learning paradigms have revealed the effectiveness of our proposed DRTL on few-shot RealSR. 
\end{itemize}
%We conduct extensive experiments based on different baseline schemes, \egno, pre-training based scheme and meta-learning based scheme. The experimental results on FS-RSR datasets with few-shot settings have demonstrated the effectiveness of our DRTL. DRTL is simple yet effective and can be used as a general FS-RSR framework that is compatible with many existing SR networks. We will release our source code upon acceptance.

The remaining parts of this paper are organized as follows: In Section~\ref{sec:related_works}, we make a comprehensive review of the related works for our Few-shot RealSR, including image super-resolution and real image restoration. In Section \ref{sec:approach}, we first introduce the background knowledge for two commonly-used few-shot learning paradigms, \ieno, pre-training, and meta-learning, and then present our approach, which is composed of our proposed distortion relation graph and distortion relation guided transfer learning. The experimental results are reported in Section \ref{sec:experiments}, where we make a thorough investigation for our DRTL and few-shot RealSR.
In Section~\ref{sec:conclusion}, We have a thorough conclusion for our proposed DRTL.
%Our codes are available to the research community at \url{http://staff.ustc.edu.cn/~chenzhibo/resources.html}.

\section{Related works}
\label{sec:related_works}
\subsection{Image Super-Resolution}
Deep learning has accelerated the development of image super-resolution techniques since the pioneer works SRCNN~\cite{dong2015image} and FSRCNN~\cite{dong2016accelerating}. Most works \cite{lim2017enhanced,qiu2019embedded,dai2019second,kim2016accurate,zhang2018image,chen2020prePIT,liang2021swinirIR,li2020mdcnTCSVT,zhang2021twoTCSVT,hu2019channelTCSVT,shi2022criteriaTCSVT,chen2022learningTCSVT} are only devoted into the synthetic degradation (\ieno, bicubic downsampling).
However, in the real-world scenario, the degradation factors are composed of hybrid distortions, such as blur, noise, and compression artifacts, etc. To tackle the challenge of real-world super-resolution, a series of works~\cite{lugmayr2019unsupervised,yoon2021simple,castillo2021generalized,umer2020deep} apply image translation technology and cycle consistence~\cite{zhu2017unpairedCycleGAN} to implement unsupervised image super-resolution, which have achieved great subjective quality. However, these methods typically suffer from unsatisfactory texture synthesis due to the lack of supervision of clean images. There are also some works~\cite{liu2022blindSR,yue2022blindSR,luo2022deepblindSR,luo2022learningblindSR,tao2021spectrumblindSR,zhang2021designingblindSR} designed to solve the blind/unknown image super-resolution, of which the degradations are still far from the real-world distortions.
Meanwhile, some real image super-resolution (RSR) datasets with limited real-world clean/distorted image pairs have been collected and released (\ieno, RealSR~\cite{cai2019toward} and DRealSR~\cite{wei2020component}), which are costly from the perspective of time and manpower. Despite some frameworks~\cite{cai2019toward,wei2020component,dong2021frequency,wei2020aimRealSR,wang2021unsupervised,li2021learning,pang2020fan,wei2020aimRealSR,xu2022dualRealSR} have achieved great progress in RealSR, they ignore the fact that limited data in real-world will prevent their further improvements. 
Different from the above works, we are the first to have a close look at the brand-new direction for RealSR, \ieno, few-shot real image super-resolution (RealSR), which is vital for the application of SR methods in the real world.

%  Pioneer works are devoted into solving the synthetic known 

% Compared with the traditional methods that are mostly based on handcrafted image priors, learning-based methods could automatically capture the statistic prior of original images from distorted images. 

\subsection{Real Image Restoration}
Collecting a large-scale clean/distorted training dataset in the real world is non-trivial, severely preventing the successful application of the fully-supervised image restoration (IR) methods. To reduce the dependencies for real-world datasets, there are three categories of schemes are proposed to improve the performance of the IR model on real-world scenarios, \ieno, unsupervised learning-based methods~\cite{ulyanov2018deepDIP,yuan2018unsupervised,du2020learningInvarintIR}, distortion simulation-based methods~\cite{wang2021realESRGAN}, and transfer learning-based methods~\cite{wei2019semi,yasarla2020syn2real,soh2020meta,park2020fastAdapt,kim2020transfer}. Unsupervised learning-based methods usually adopt unpaired distorted/clean images or only distorted images for optimization with generative models.  
For instance, Ulyanov~\etal~\cite{ulyanov2018deepDIP} propose to utilize the CNN to capture the deep image statistics (named as the deep image prior, DIP) through an iterative self-supervised optimization. Yuan~\etal~\cite{yuan2018unsupervised} introduces a Cycle-in-Cycle structure to exploit unpaired distorted/clean images for more general image super-resolution. Du~\etal~\cite{du2020learningInvarintIR} introduces a discrete disentangling representation learning method to capture the invariant clean representations from unpaired distorted/clean image pairs. Distortion simulation-based methods~\cite{wang2021realESRGAN} aim to simulate real-world distortions and utilize the synthesized datasets to optimize the IR network, which achieves excellent performance in real-world scenarios.  The final category of IR methods intends to leverage transfer learning techniques to achieve image restoration for real distorted images. For example, Wei~\etal~\cite{wei2019semi,yasarla2020syn2real} propose to capture the distortion priors of rain streaks from the auxiliary synthetic/fake rain streaks to achieve the clean image restoration in a semi-supervised learning manner. However, the above methods are designed based on the assumption that rain streaks can be modeled with Gaussian distribution, which is not always satisfied for other distortions. With the advancement of transfer learning, Soh~\etal~\cite{soh2020meta} and Park~\etal~\cite{park2020fastAdapt} propose to leverage the Meta-transfer learning to deal with the challenging task of zero-shot super-resolution (ZSSR). Kim~\etal~\cite{kim2020transfer} utilize the adaptive instance normalization to realize the knowledge transfer from synthetic noise to real noise. Different from the above works, investigating the knowledge transfer between the homogeneous distortions (\egno,  noise to noise, rain to rain, etc.), our DRTL focus on a more challenging setting, where the auxiliary tasks can be heterogeneous synthetic distortions (\egno, the jpeg artifacts to RealSR, mixed distortions to RealSR, etc.).
\begin{figure*}
	\centering
	\includegraphics[width=\textwidth]{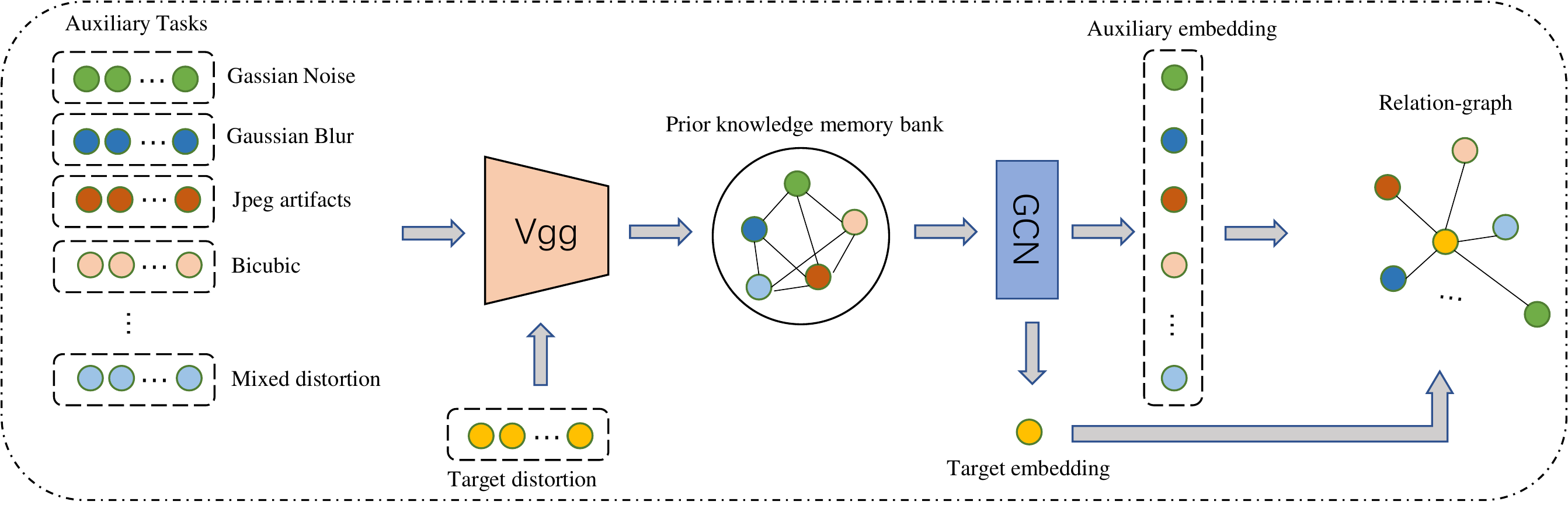}
	\caption{Architecture of our proposed distortion relation network (DRN), which consists of feature extractor(VGG11), prior knowledge memory bank, and Graph convolution network. The representation of each distortion sample can be tapped into the prior knowledge memory bank to capture the distortion embedding with GCN as the node of the distortion relation graph. Then the edges of the distortion relation graph can be computed with the cosine similarity of different nodes.}
	\label{fig:framework_graph}
\end{figure*}
\section{Approach}
\label{sec:approach}
\subsection{Recap of Transfer Learning}
In this section, we clarify two commonly-used transfer learning methods, \ieno, pre-training, and meta-learning, as the basis of our proposed distortion relation guided transfer learning (DRTL). 

\subsubsection{Pre-training based Transfer Learning}
As the basic transfer learning technology, pre-training has been widely applied to different vision tasks \cite{chen2020prePIT,he2019rethinking}. Pre-training-based transfer learning can be divided into two processes, including pre-training and fine-tuning. Given the auxiliary tasks $\mathcal{T}^{a}$, target task $\mathcal{T}^t$ and
%pre-training based methods~\cite{} typically first train models on $\mathcal{T}^{au}$, and then directly fine-tune the model on $\mathcal{T}^t$. 
a learning model $f_\theta$ with parameters $\theta$, the pre-training process  aims to learn the optimal model parameters on auxiliary tasks $\mathcal{T}^{a}$ as Eq.~\ref{equ:optim} to obtain the task-relevant knowledge,
\begin{equation}
      \underset{\theta}{\mathop{\min}}\,\sum\limits_{{{\mathcal{T}_{i}}^{a}\sim p({{\mathcal{T}^{a}})}}}{{\mathcal{L}_{\mathcal{T}_{i}^{a}}}({{f}_{\theta }(x_i^a), y_i^a})}.
      \label{equ:optim}
\end{equation}
% \begin{equation}
%      {\theta_i}'=\theta -\alpha {{\nabla }_{\theta }}{{\mathcal{L}}_{\mathcal{T}_{i}^{a}}}({{f}_{\theta }}(x_i^a), y_i^a),
%      \label{equ:update}
% \end{equation}
where the ${{\mathcal{L}}_{\mathcal{T}_{i}^{a}}}$ refers to the optimization loss in the $i^{th}$ auxiliary task $\mathcal{T}_{i}^{a}$, ${{x}_i^a}$ and ${{y}_i^a}$ denotes the samples and their corresponding labels of the task $\mathcal{T}_{i}^{a}$.
% as Eq.~\ref{equ:loss}
% \begin{equation}
%       {{\mathcal{L}_{\mathcal{T}_{i}^{a}}}({{f}_{\theta}})=\sum\limits_{{{x}^{(j)}},{{y}^{(j)}}\in \mathcal{T}_{i}^{a}}{||{{f}_{\theta }}({{x}^{(j)}}),{{y}^{(j)}}||}}.
%       \label{equ:loss}
% \end{equation}
% Here, ${{x}^{(j)}}$ and ${{y}^{(j)}}$ are $j^{th}$ distorted/clean image pairs in task $\mathcal{T}_{i}^{a}$.
 The optimization objective of pre-training is to minimize the loss function in all auxiliary tasks.

In the fine-tuning stage, the best parameters $\theta_m$ in the pre-training stage are used as the initial parameters. Then model $f_{\theta_m}$ is updated with the optimization objective to acquire the best performance in the target task as Eq.~\ref{equ:fine_stage}:
\begin{equation}
    \underset{\theta }{\mathop{\min }}\,{{\mathcal{L}}_{{{\mathcal{T}}^{t}}}}({{f}_{\theta_m}}(x^t), y^t),
    \label{equ:fine_stage}
\end{equation}
where $x^t$ and $y^t$ are the training samples and their corresponding labels of the target task.

\subsubsection{Meta-learning based Transfer Learning}
Different from pre-training-based transfer-learning, which directly learns the task-relevant knowledge from auxiliary tasks. Meta-learning-based Transfer Learning aims to learn the capability of fast adaptation to all auxiliary tasks \cite{finn2017modelMAML}. The typical work is MAML~\cite{finn2017modelMAML} (Model-agnostic Meta-learning), which can be divided into two processes, respectively as Meta-Train and Meta-Test. For the Meta-Train process, the model is first optimized with multiple tasks $\{\mathcal{T}_i^a \in \mathcal{T}^a, 1 \le i \le N\}$ as Eq.~\ref{equ:update}
\begin{equation}
     {\theta_i'}=\theta -\alpha {{\nabla }_{\theta }}{{\mathcal{L}}_{\mathcal{T}_{i}^{a}}}({{f}_{\theta }}(x_i^a), y_i^a),
     \label{equ:update}
\end{equation}
%${{\mathcal{L}}_{\mathcal{T}_{i}^{a}}}$, that is randomly sampled from auxiliary tasks $p({\mathcal{T}}^{a})$ as Eq.~\ref{equ:update}. 
To enable the model to have a fast adaptation capability with few-shot samples, the model is optimized through the second-order gradient from the multiple tasks of $\mathcal{T}^a$ based on the look-ahead gradient $\theta_i'$ in Eq.~\ref{equ:update} as:
\begin{equation}
\theta_m = \theta - \beta {{\nabla }_{\theta }}\sum\limits_{\mathcal{T}_{i}^{a}\sim{\ }p({{\mathcal{T}}^{a}})}{{{\mathcal{L}}_{\mathcal{T}_{i}^{a}}}(f_{\theta_i'}(x_i^a), y_i^a)}.
\end{equation}
% As described in Eq.~\ref{equ:meta_optim}, Meta-Train aims to learn the general $\theta$, which can be transferred to all tasks well. 
The Meta-Test stage is the same as fine-tuning, which can be represented as Eq.~\ref{equ:fine_stage}.
\begin{figure*}[ht]
	\centering
	\includegraphics[width=0.95\textwidth]{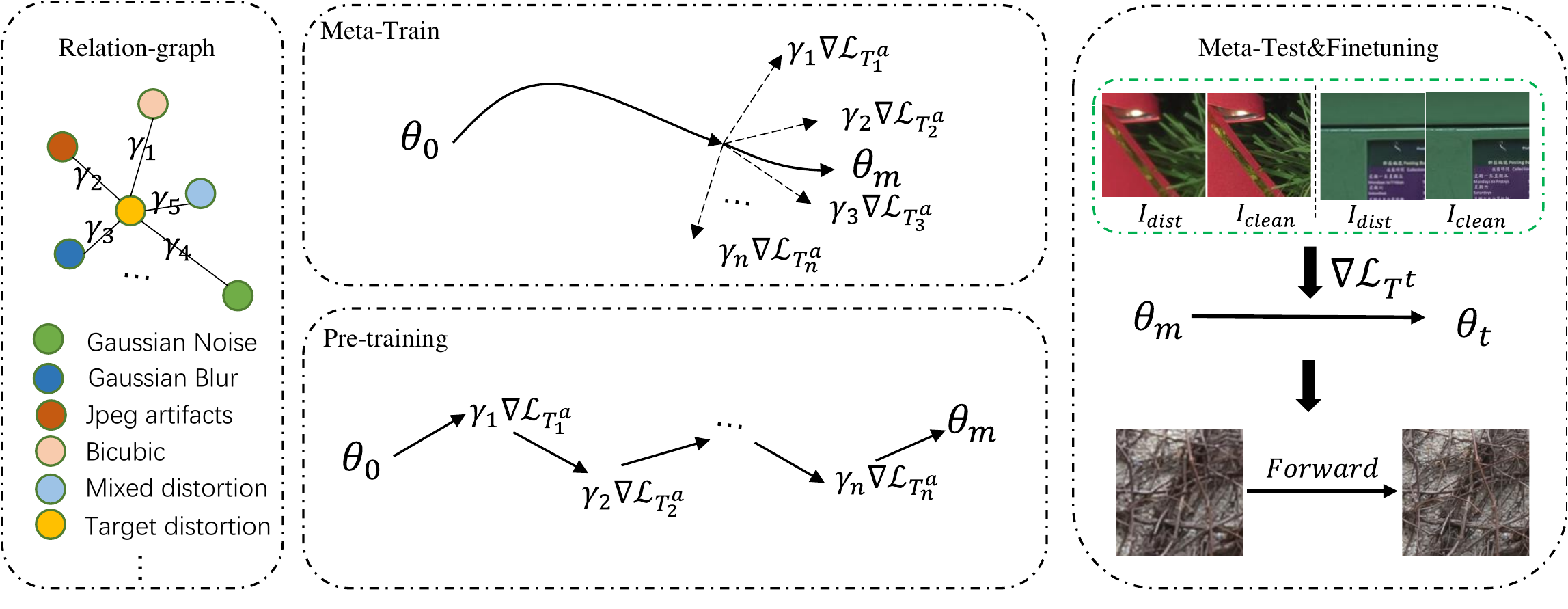}
	\caption{Overall workflow of Distortion Relation guided Transfer-Learning (DRTL). $I_{dist}$ and $I_{clean}$ denotes the distorted and high-quality image pairs in RealSR.}
	\label{fig:framework_optimization}
\end{figure*}

\subsection{Building Distortion Relation Graph}
The distortion relation graph aims to explore and measure the transferability between multiple synthetic distortions and target real-world distortion. Once getting such a relation graph, 
the most transferable auxiliary distortions can be selected to assist the optimization of target real SR with few-shot samples. Moreover, we could leverage the distortion relation to guide the knowledge transfer process from auxiliary distortions to the target RealSR. In this paper, we propose a  distortion 
relation graph, where the nodes in the graph denote the relation embeddings between each sample and prior knowledge memory bank. The edges are created based on the similarity between different nodes corresponding to different distortions (\ieno, the transferability between different distortions). The architecture of this distortion relation graph is described in Figure \ref{fig:framework_graph} and the graph is generated with our proposed distortion relation network (DRN), which consists of four key components: distortion-aware feature extractor, prior knowledge memory bank, relation-aware nodes, and edges. We will introduce each component in detail in the following sub-sections.

\subsubsection{Distortion-aware Feature Extractor}
%%%----失真图像通常包含有大量的原始图像先验，会干扰到失真的特征提取。一般的特征提取较难完全提取到其中的失真信息，
To reduce the interference from the texture and structure of distorted image to extracting the distortion features, we compute the residual $I_{r}$ between clean image $I$ and distorted image $I_{d}$ as ${{I}_{r}}=I-{{I}_{d}}$ as the inputs of distortion-aware feature extractor. Notably, previous works~\cite{zhang2018unreasonable,ding2020image} have demonstrated the effectiveness of VGG~\cite{simonyan2014very} network for the perception of distortions. Inspired by this, we 
%%加一些IQA的相关参考文献。
% Recently, the feature extractor of VGG have been utilized in distortion perception []. Inspired by this success, 
propose to utilize VGG-11~\cite{simonyan2014very} as our distortion-aware feature extractor to extract the distortion representation for each distortion as ${F}_{d} = \mathrm{VGG}({I}_{r})$, and the feature extractor will be optimized with the distortion classification.

\subsubsection{Knowledge Memory Bank}
Generally, graph learning aims to capture the internal relationship between samples or structures, and then assist the current task~\cite{kipf2016semi,hamilton2017inductive}. To capture the distortion relation with the graph, 
%achieve our purpose that leveraging the relation graph to re-weight/guide the knowledge transfer, 
both the target distortion and auxiliary synthetic distortions must be taken into the same graph as nodes and infer their relation weights (\ieno, the edge of graph). However, directly predicting relation weights between synthetic and real distortions will inevitably face two essential issues: First, the target distortion cannot be achieved before the establishment of a distortion relation graph. Second, directly establishing the relation graph with the current real target distortion may make the graph hard to generalize well to other unseen real distortions. To obtain a universal and general distortion relation graph, we assign a long-term distortion memory bank to record the knowledge from previously seen distortion types, where the distortion knowledge is stored with a memory graph $\mathcal{M}=(\mathcal{H}_\mathcal{M}, \mathcal{A}_\mathcal{M})$. Here, the memory nodes $\mathcal{H}_m$ and edges $\mathcal{A}_\mathcal{M})$ are trainable in the  optimization process of our distortion relation network (DRN). 
   % Based on the distortion memory bank, the knowledge and relation of auxiliary synthetic distortions can be automatically established and stored in the distortion memory bank as the training goes on. 
   In this way, each distortion can be tapped into the  distortion memory bank to obtain its corresponding representation/node in the same space, which is suitable for measuring the distortion relation.
 Note that, with the distortion memory bank, the target distortion and auxiliary distortions can extract their corresponding nodes independently.  
\begin{algorithm}
\caption{Distortion Relation Graph}
\label{alg:1}
\begin{algorithmic}[1]
\State \textbf{Inputs:} Auxiliary distortions $\mathcal{T}_{i}^a$, where $1\le i \le N$ and $N$ represents the number of the auxiliary distortions; Target distortion $\mathcal{T}^{t}$ (\ieno, RealSR); Our proposed distortion relation network (DRN); 
\State Get the distortion embedding nodes $c_a^i$ for each auxiliary distortion with 
$c_a^i = \mathrm{DRN}(\mathcal{T}_{i}^a)$.
\State Get the distortion embedding nodes $c_t$ for target distortion with $c_t= \mathrm{DRN}(\mathcal{T}^{t})$.
\State Compute the relation of the $i$-th auxiliary distortion and target distortion with 
$\gamma_i = cosine\_simlarity(c_a^i, c_t)$
\State \textbf{Output:} The relation matrix $\{\gamma_i | 1 \le i \le N \}$
\end{algorithmic}
\end{algorithm}

\begin{algorithm}
\caption{The Pre-training Stage of DRTL$_p$}
\label{alg:2}
\begin{algorithmic}[1]
\State \textbf{Inputs:} Auxiliary distortions $\mathcal{T}_i^a$, where $1 \le i \le N$; Target distortion $\mathcal{T}^{t}$; Relation matrix  $S = \{\gamma_i | 1 \le i \le N \}$; A restoration model $f_\theta$;  learning rate: $\alpha$.
\For{iteration = 0  to $T$}
\For{i = 0 to $N$}
 \State Update $\theta$ with ${\theta_i'}=\theta -\alpha {{\nabla }_{\theta }}{\gamma_i}{{\mathcal{L}}_{\mathcal{T}_{i}^a}}({{f}_{\theta }})$;
\EndFor
\EndFor
\State \textbf{Output:} The optimal initial restoration network ${f_\theta}_m$ on target distortion.
\end{algorithmic}
\end{algorithm}

\subsubsection{Graph Nodes and Edges.} 
Our distortion relation graph $\mathcal{R}=(\mathcal{C}_\mathcal{R}, \mathcal{A}_\mathcal{R})$ consists of two essential components, \ieno, Nodes ${{\mathcal{C}}_{\mathcal{R}}}=\{{{c}^{i}}|\forall i\in [1,K]\}\in {{\mathbb{R}}^{K\times d}}$ and Edges ${{\mathcal{A}}_{\mathcal{R}}}=\{|{{\mathcal{A}}_{\mathcal{R}}}({{c}^{i}},{{c}^{j}})|\forall i,j\in [1,K]\}\in {{\mathbb{R}}^{K\times K}}$, where $k$ and $d$ are the number and dimension of nodes. As stated in the above section, the nodes of the distortion relation graph denote the feature representation of each distortion, which is obtained by projecting distorted samples to the knowledge memory bank $\mathcal{M}=(\mathcal{H}_\mathcal{M}, \mathcal{A}_\mathcal{M})$. To generate the nodes of distortions, we require to warp the features of different distortions into the same space with the memory bank. Concretely, we 
first compute the distance between the $i^{th}$ distortion feature $F^{i}_{d}$ and each node of the memory bank as Eq.~\ref{equ:relation}:
\begin{equation}
     \mathcal{A}_{p}^{i}=\sigma (||{{F^{i}_{d}}}-{{\mathcal{H}}_{\mathcal{M}}}||_{2}^{2}),
     \label{equ:relation}
\end{equation}
where $\sigma$ is a linear transform. Then we can warp the distortion feature into the space of the memory bank with graph convolution network (GCN)~\cite{kipf2016semi}. In the computing process, the adjacent matrix contains two parts, including the edges of memory bank  ${{\mathcal{A}}_{\mathcal{M}}}$ and the distance matrix of each distortion with memory nodes $\mathcal{A}_{p}^{i}$ as: 
\begin{equation}
    \mathcal{A}=[\mathcal{A}_{p}^{i},{{\mathcal{A}}_{\mathcal{M}}}],
    \label{equ:edgecat}
\end{equation}
where $\mathcal{A}_\mathcal{M}$ can be computed by measuring the distances between different memory nodes as:
\begin{equation}
    {{\mathcal{A}}_{\mathcal{M}}}=\{\sigma (||{{\mathcal{H}}_{m}}-{{\mathcal{H}}_{n}}||_{2}^{2})|m,n\in \{1,Q\}\}.
    \label{equ:bank_adj}
\end{equation}
In this way, we can obtain the nodes of each distortion in the memory space with GCN~\cite{kipf2016semi} as:
\begin{equation}
    c_i = \mathrm{GCN}([{{F}^{i}_{d}}, \mathcal{H}_\mathcal{M}], \mathcal{A}).
    \label{equ:gcn}
\end{equation}
After obtaining the nodes ${{\mathcal{C}}_{\mathcal{R}}}=\{{{c}^{i}}|\forall i\in [1,K]\}\in {{\mathbb{R}}^{K\times d}}$ for distortion relation graph, we utilize cosine similarity to measure the distance of different distortion nodes, which acting as the edges ${{\mathcal{A}}_{\mathcal{R}}}$ of the distortion relation graph and is used to measure the transferability between different distortions. Based on the above methods, we can obtain the relation $\gamma$ between each auxiliary distortion and target distortion as Alg.~\ref{alg:1}.

\subsection{Distortion Relation guided Transfer-Learning} 
To transfer the knowledge from auxiliary distortions to the target RealSR with few-shot distorted/clean image pairs, one naive method is to utilize the classic transfer learning technology to achieve Few-shot RealSR. However, directly applying transfer learning ignores the relation between auxiliary distortions and target RealSR, and thus is easy to cause the sub-optimal solution.
In order to extract reliable and optimal distortion knowledge from auxiliary synthetic distortions to target RealSR,  in this paper, we propose distortion relation guided transfer-learning (DRTL). The workflow of DRTL is described in Figure~\ref{fig:framework_optimization}, which is based on two intuitions: 1) multiple auxiliary distortions that are similar to the target RealSR can provide more transferable knowledge for RealSR than single distortion. 2) When the auxiliary distortion is more relevant to the target RealSR, the knowledge transferability will be higher. Therefore, we should increase the proportion of this distortion knowledge in the optimization of the Pre-training/Meta-Train process.
% \begin{table*}[htp]
% \centering
% \caption{The way of auxiliary distortions generation. To save space, the distortion type have been abbreviated as follows: Bicubic downsampling (i.e., Bicubic), Bicubic downsampling with anistropic blurring (i.e., Ani\_bic), Gassian noise (i.e., Noise), Gaussian Blur (i.e., Blur), Mixed mild (i.e., Mild), Mixed moderate (i.e., Moderate), Mixed severe (i.e., Severe).}
% \label{tab:aux}
% \begin{tabular}{l|l|l}
% \hline
% Distortion types & \multicolumn{2}{l}{Generation} \\ \hline
% Bicubic & \multicolumn{2}{l}{Bicubic with scale 8} \\ \hline
% Ani\_bic & \multicolumn{2}{l}{Bicbic with scale 4 + anistropic blur} \\ \hline
% Noise & \multicolumn{2}{l}{Gassian noise with $\sigma$ from the range of [0, 50]} \\ \hline
% Blur & \multicolumn{2}{l}{Gaussin blur with  $\sigma$ from the range of [0, 5]} \\ \hline
% Mild & \multicolumn{2}{l}{Gaussian noise + Gaussian blur + JPEG artifacts; Distortion level lies in the range of [9, 11]} \\ \hline
% Moderate & \multicolumn{2}{l}{Gaussian noise + Gaussian blur + JPEG artifacts; Distortion level lies in the range of [12, 17]} \\ \hline
% Severe & \multicolumn{2}{l}{Gaussian noise + Gaussian blur + JPEG artifacts; Distortion level lies in the range of [18, 20]} \\ \hline
% \end{tabular}
% \end{table*}

%Since sufficient and comprehensive relevant auxiliary tasks could further bring the performance improvements for the target real super-resolution, 

\begin{algorithm}
\caption{The Meta-Train Stage of DRTL$_m$}
\label{alg:3}
\begin{algorithmic}[1]
\State \textbf{Inputs:} Auxiliary distortions $\mathcal{T}_i^a$, where $1 \le i \le N$; Target distortion $\mathcal{T}^t$; Relation matrix  $S = \{\gamma_i | 1 \le i \le N \}$; learning rate: $\alpha$ and $\beta$.
\For{iteration = 0  to $T$}
\For{i = 0 to $N$}
\State Sample batch from auxiliary distortions $\mathcal{T}_{i}^a$;
\State Evaluate gradient ${{\nabla }_{\theta }}{{\mathcal{L}}_{\mathcal{T}_{i}^a}}({{f}_{\theta }})$ with respect to sampled batch of distortion $\mathcal{T}_{i}^a$;
\State Update parameters: ${\theta_i'}=\theta -\alpha {{\nabla }_{\theta }}{{\mathcal{L}}_{\mathcal{T}_{i}^a}}({{f}_{\theta }})$;
\EndFor
\State Update parameters: $\theta \leftarrow \theta -\beta {{\nabla }_{\theta }}\sum\limits_{i=1}^{K}{\gamma_i}{{{\mathcal{L}}_{{{\mathcal{T}}_i^a}}}(f_{\theta_i'})}$
\EndFor
\State \textbf{Output:} The optimal initial model ${f_\theta}_m$  on target distortion.
\end{algorithmic}
\end{algorithm}
To achieve this, our proposed DRTL aims to utilize the distortion relation to guide the optimization process of each auxiliary distortion in the transfer learning. The DRTL can be divided into three steps: distortion relation computing, Pre-training/Meta-Train with auxiliary distortions, and Finetuning/Meta-Test with target RealSR. 
\begin{figure*}
	\centering
\includegraphics[width=0.85\linewidth]{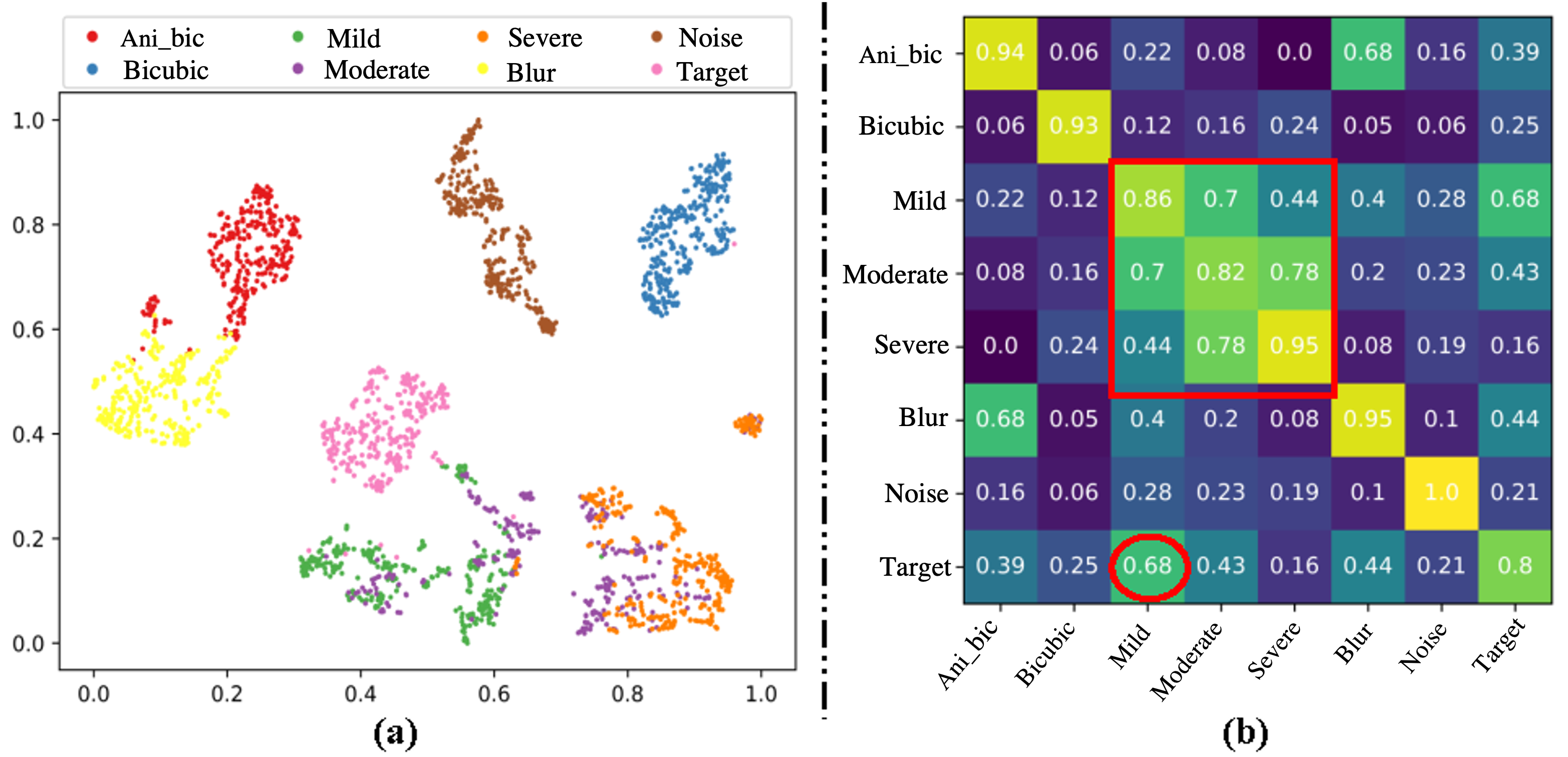}
	\caption{(a) Visualization of nodes in the distortion relation graph and (b) Visualization of the adjacent matrix in the distortion relation graph. Target refers to the distortion in RealSR.}
	\label{fig:cluster}
\end{figure*}
As shown in Figure~\ref{fig:framework_optimization}, in the first step, we can obtain the distortion relation $\gamma$ by computing the edges of our pre-trained distortion relation graph. Here, the $\gamma$ will increase when the distortions are more relevant. In the second step, we utilize the relation coefficient $\gamma$ to guide the transfer learning process in a gradient modulation/re-weighting manner. In this paper, we instantiate our DRTL with two commonly-used transfer learning techniques, including Pre-training, and MAML, as DRTL$_p$ and DRTL$_m$, respectively. For DRTL$_p$, we revise the optimization process of pre-training in Eq.~\ref{equ:optim_pretraining} as:
\begin{equation}
      % \underset{\theta}{\mathop{\min}}\,
      \theta_m = \theta -  \alpha {{\nabla }_{\theta }
      \sum\limits_{{{\mathcal{T}_{i}}^{a}\sim }p({{\mathcal{T}^{a}})}}}{{\gamma_i \mathcal{L}_{\mathcal{T}_{i}^{a}}}({{f}_{\theta }(I_{d_i}^a), I_i^a})},
      \label{equ:optim_pretraining}
\end{equation}
where $I_{d_i}^a$ and $I_i^a$ represent the distorted and clean images in the $i^{th}$ auxiliary distortion task.
For DRTL$_m$, we revise the optimization process of Meta-Train in Eq.~\ref{equ:meta_optim} as:
\begin{equation}
\theta_m = \theta - \beta {{\nabla }_{\theta }}\sum\limits_{\mathcal{T}_{i}^{a}\sim{\ }p({{\mathcal{T}}^{a}})}{{{\gamma_i \mathcal{L}}_{\mathcal{T}_{i}^{a}}}(f_{\theta_i'}(I_{d_i}^a), I_i^a)}.
\label{equ:meta_optim}
\end{equation}
\begin{table}[htp]
\centering
\caption{The way of auxiliary distortions generation. To save space, the distortion type have been abbreviated as follows: Bicubic downsampling (i.e., Bicubic), Bicubic downsampling with anistropic blurring (i.e., Ani\_bic), Gassian noise (i.e., Noise), Gaussian Blur (i.e., Blur), Mixed mild (i.e., Mild), Mixed moderate (i.e., Moderate), Mixed severe (i.e., Severe).}
\label{tab:aux}
\begin{tabular}{c|c}
\hline
Distortion types & Generation                                    \\ \hline
Bicubic          & Bicubic with scale 8                           \\ \hline
Ani\_bic          & Bicbic with scale 4 + anistropic blur         \\ \hline
Noise            & Gassian noise with $\sigma$ from the range of [0, 50] \\ \hline
Blur             & Gaussin blur with $\sigma$ from the range of [0, 5]\\ \hline
Mild             & \begin{tabular}[c]{@{}c@{}}Gaussian noise + Gaussian blur + JPEG artifacts;\\ Distortion level lies in the range of [9, 11]\end{tabular}   \\ \hline
Moderate         & \begin{tabular}[c]{@{}c@{}}Gaussian noise + Gaussian blur + JPEG artifacts;\\ Distortion level lies in the range of [12, 17]\end{tabular} \\ \hline
Severe           & \begin{tabular}[c]{@{}c@{}}Gaussian noise + Gaussian blur + JPEG artifacts;\\ Distortion level lies in the range of [18, 20]\end{tabular}  \\ \hline
\end{tabular}
\end{table}
In this way, the optimization direction of MAML and pre-training can be closer to target distortion, which can capture most task-relevant knowledge.
In the process of Meta-Test or fine-tuning, the optimal transferable model parameters $\theta_m$ only need to be fine-tuned with the few-shot distorted/clean pairs of target distortion. The algorithms of the pre-training stage in DRTL$_p$  and Meta-Train stage of DRTL$_m$ are shown in Alg.~\ref{alg:2} and~\ref{alg:3}.

%Since the performance of transfer learning is influenced by the relevance of auxiliary tasks and target task, 

\begin{table*}[htp]
\centering
\caption{Quantitative comparisons of our proposed DRTL, Pre-training, MAML and Baseline on testing dataset of target distortion. Baseline methods refer to  directly training with few-shot clean-distorted image pairs without transfer learning. DRTL(PT) denotes the DRTL with pre-training.}
\begin{tabular}{c|ll|ll|ll|ll|ll}
% \toprule
\hline
\multirow{2}{*}{Models} & \multicolumn{2}{c|}{Baseline} & \multicolumn{2}{c|}{Pre-training} & \multicolumn{2}{c|}{MAML} & \multicolumn{2}{c|}{DRTL$_p$} & \multicolumn{2}{c}{DRTL$_m$} \\ \cline{2-11} 
 & PSNR & SSIM & PSNR & SSIM & PSNR & SSIM & PSNR & SSIM & PSNR & SSIM \\ \hline%\midrule
DnCNN \cite{zhang2017beyond} & 29.297 & 0.8738 & 30.858 & 0.8861 & 23.466 & 0.8060 & 30.977 & 0.8892 & \textbf{31.101} & \textbf{0.8908} \\
VDSR \cite{kim2016accurate} & 31.067 & 0.8849 & 31.290 & 0.8901 & 31.219 & 0.8877 & 31.358 & \textbf{0.8908} & \textbf{31.367} & 0.8907 \\ 
RCAN \cite{zhang2018image} & 31.299 & 0.8933 & 31.698 & 0.8977 & 31.546 & 0.8946 & \textbf{31.811} & \textbf{0.8988} & 31.597 & 0.8956 \\ 
RDN \cite{zhang2018residual} & \multicolumn{1}{l}{31.235} & \multicolumn{1}{l|}{0.8914} & \multicolumn{1}{l}{31.535} & \multicolumn{1}{l|}{0.8948} & \multicolumn{1}{l}{31.392} & \multicolumn{1}{l|}{0.8926} & \multicolumn{1}{l}{\textbf{31.655}} & \multicolumn{1}{l|}{\textbf{0.8965}} & \multicolumn{1}{l}{31.493} & \multicolumn{1}{l}{0.8940} \\ 
SwinIR~\cite{liang2021swinirIR} & \multicolumn{1}{l}{31.336} & \multicolumn{1}{l|}{0.8897} &  \multicolumn{1}{l}{31.526} & \multicolumn{1}{l|}{0.8956}  & \multicolumn{1}{l}{31.402} & \multicolumn{1}{l|}{0.8923} &   \multicolumn{1}{l}{\textbf{31.785}} & \multicolumn{1}{l|}{\textbf{0.8986}} & \multicolumn{1}{l}{31.447} & \multicolumn{1}{l}{0.8924} \\
\hline%\bottomrule
\end{tabular}
\label{tab:SOTA}
\end{table*}
\section{Experiments}
\label{sec:experiments}
In this section, we first introduce the auxiliary distortions datasets and target RealSR datasets in Sec.~\ref{sec:datasets}. Then, we present the implementation details of our DRTL in Sec.~\ref{sec:details}. And, we show a comprehensive comparison and analysis of the components in DRTL to demonstrate their effectiveness and superiority in Sec.~\ref{sec:vis}, Sec.~\ref{sec:graph}, Sec.~\ref{sec:sota}, Sec.~\ref{sec:ablation}.

\subsection{Datasets}
\label{sec:datasets}

\noindent\textbf{Auxiliary datasets.}
To demonstrate the effectiveness and robustness of our DRTL, \textit{we select most relevant 7 common synthetic distortions as auxiliary tasks based on our proposed distortion relation graph}, including Bicubic downsampling \cite{agustsson2017ntire},  Bicubic downsampling with An-isotropic kernels \cite{soh2020meta}, Gaussian noise, Gaussian blur, Mixed mild distortion, Mixed moderate distortion, Mixed severe distortion \cite{suganuma2019attention}. Then, we take 800 clean images of the DIV2K dataset~\cite{agustsson2017ntire} as original clean images and add the above-mentioned distortions into the clean images to generate 7 auxiliary datasets. 
The way of distortion generation is shown in Table \ref{tab:aux}. Particularly, the mixed distortions~\cite{yu2018crafting,li2020learningFDRNet} are composed of Gaussian noise, Gaussian blur, and compression artifacts, which are divided into 10 levels with $\sigma \sim$ [0, 50], $\sigma \sim$ [0, 5] and compression quality $q \sim$ [10, 100], respectively. Following the \cite{yu2018crafting}, the mixed distortion can be divided into three levels, \ieno, mild, moderate, and severe. Mixed mild distortion denotes the total distortion level of Gaussian noise, Gaussian blur, and Jpeg artifacts in the range of [9,11]. Mixed moderate distortion and mixed severe distortion are in the range of [12,17] and [18,20], respectively. 

\noindent\textbf{Target dataset}
 We select the typical RealSR~\cite{cai2019toward} dataset as our target task. To get a few-shot real-world dataset for training/fine-tuning, we randomly select 30 clean-distorted image pairs from the train set of RealSR~\cite{cai2019toward} as our training dataset with target distortion. For the final evaluation, we further use 30 clean/distorted images in the test set of RealSR~\cite{cai2019toward} as our final testing dataset. Note that, the target clean-distorted image pairs have the same resolution. The distorted images contain complicated real-world distortions, which are captured by Canon and Nikon cameras~\cite{cai2019toward}.

% 数据集的合成方式，包含14种常见失真，目标失真任务的介绍。
\subsection{Implementation Details}~\label{sec:details}
%Since the proposed DRTL is a model-agnostic optimization strategy, we adapt the general model VDSR~\cite{kim2016accurate}, that contains a series of non-linear Resblocks, as our baseline model. 
The implementation of DRTL is based on the PyTorch framework. The whole training process can be divided into two stages: 1) Pre-training/Meta-Train. 2) Fin-tuning/Meta-Test. For the first step, we utilize Adam optimizer with an initial learning rate of 0.0001 for optimization. For the Fintuning/Meta-Test step, the learning rate decay by a factor of 0.8 every 3000 iterations. We set the batch size as 32 and leverage random flip, rotation, and cropping to achieve data augmentation. The size of cropped image is 64$\times$64. $L_1$ loss has been proved effective to optimize the model, especially in image restoration \cite{zhang2018image,cai2019toward}. Therefore, we only use $L_1$ loss to optimize the DRTL in this paper. For the distortion relation graph, we optimize the distortion relation network to have the capability for identifying distortions and capturing their relation with the distortion classification constraint.  

% \begin{figure}
% 	\centering
% 	\includegraphics[width=0.55\linewidth]{fig/relation_all.pdf}
% 	\caption{Visualization of adjacent matrix in distortion relation graph.}
% 	\label{fig:adjacent}
% 	\vspace{-6mm}
% \end{figure}

% few-shot定义的解释，auxiliary池的构建等
% no pre-training (baseline), Pre-training, Meta-learning, Deep image prior, Learning invariant
%------results: （30张图）
%%%%%%%% || directly training  || pre-training || Meta-learning || Ours || DIP || Learning invriant
\begin{figure}
	\centering
	\includegraphics[width=1.0\linewidth]{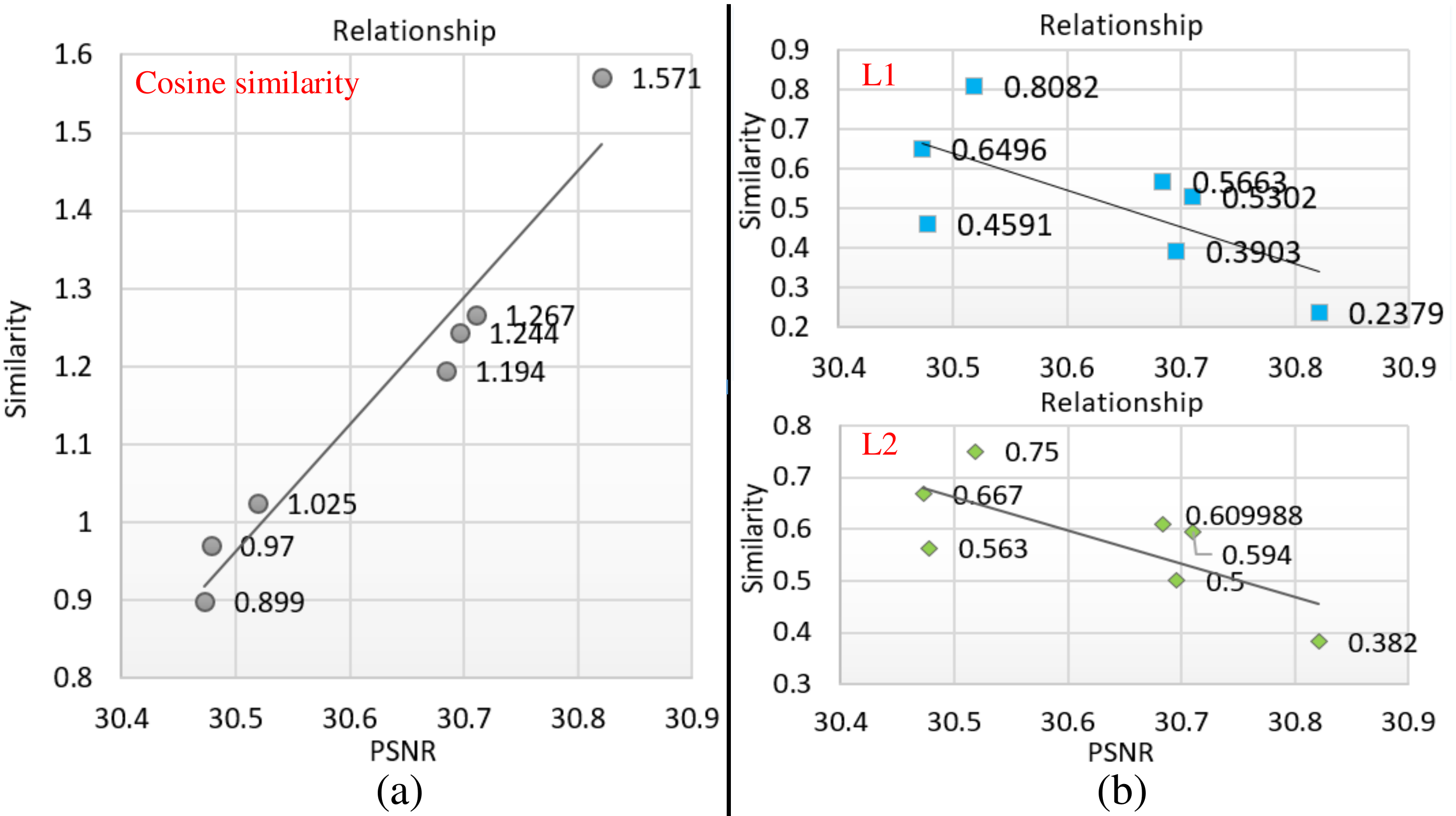}
	\caption{Visualization of the relationship between similarity of distortions and transfer-ability. Here, the transfer-ability is measured with PSNR. (a) Modeling the similarity with cosine similarity, which is adopted in our paper. (b) Modeling the similarity with L1 or L2 loss.}
	\label{fig:Relation}
\end{figure}
\subsection{Graph Visualization and Explanation}~\label{sec:vis}
In this section, we elaborately discuss how our distortion relation graph works. As shown in Figure~\ref{fig:cluster}(a), we visualize the feature embedding for each sample with the auxiliary and target distortions. With the distortion relation network (DRN), the features of all samples with the same distortion have been clustered in the same region/cluster, which reveals that DRN could successfully capture/model the characteristics of each distortion. It is noteworthy that, despite the target distortion not being seen during the DRN training, it still can be clustered/categorized well in the feature space. 
\begin{figure*}
	\centering
	\includegraphics[width=\textwidth]{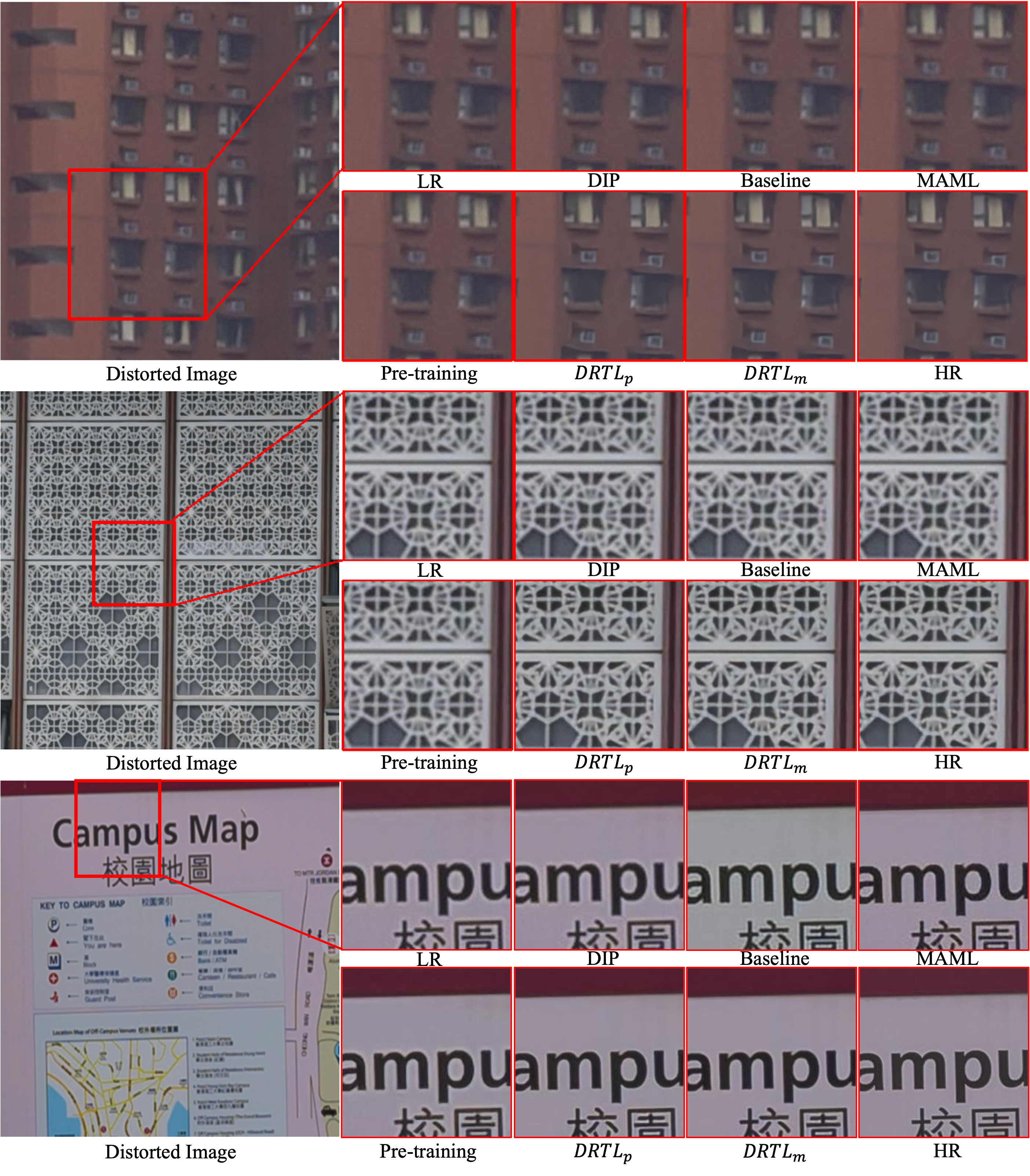}	\caption{Subjective comparison of our DRTL with the state-of-the-art methods. Here, DRTL$_p$ and DRTL$_m$ denote our DRTL is integrated into Pre-training and MAML.}
\label{fig:subjective}
\end{figure*}
To better understand the learned relation between different auxiliary distortions and target distortion (\ieno, RealSR), we further visualize the adjacent similarity matrix of different distortion nodes learned by our distortion relation graph in Figure~\ref{fig:cluster}(b). According to this adjacent matrix, we can see that the target distortion is more similar (with a similarity of 0.68) to the mixed mild distortion. Moreover, we observe that the mixed mild, moderate, and severe distortions are similar/closed since these distortions contain the three same distortion types with different levels, which reveals our proposed distortion relation graph indeed could effectively capture/model the relation of different kinds of distortions.

\subsection{Effectiveness of Leveraging Distortion Relation}~\label{sec:graph}
To validate the correlation between the distortion relation and the transferable knowledge in the auxiliary distortions, we measure the transferability of each auxiliary distortion by utilizing each auxiliary distortion as the pre-training task individually. Then we can measure its corresponding transferability with the performance of the target RealSR (\ieno, PSNR). The experimental results are shown in Figure~\ref{fig:Relation}(a), where we can have the conclusion: with the increase of distortion relation, the  auxiliary distortion contains more powerful transferable knowledge for the target SR. This also reveals the effectiveness of our distortion relation graph for the measurement of transferability. 

\subsection{Comparison with State-of-the-arts}
\label{sec:sota}
\subsubsection{Comparison with Transfer Learning based Method} 
In this section, we compare our proposed DRTL with the two transfer learning-based  methods in Table. \ref{tab:SOTA}, including the Pre-training~\cite{he2019rethinking} and MAML~\cite{finn2017modelMAML} schemes. The baseline denotes that the model is directly trained on the few-shot target distortion dataset. Since our proposed DRTL is a model-agnostic optimization strategy, we select five general models, including (DnCNN~\cite{zhang2017beyond}, VDSR~\cite{kim2016accurate}, RCAN~\cite{zhang2018image}, RDN~\cite{zhang2018residual}, and  SwinIR~\cite{liang2021swinirIR}) as backbones for evaluation. 
Specifically, we compare the training schemes of baseline, Pre-training, MAML, Pre-training based DRTL (\ieno, DRTL$_p$), and MAML-based DRTL (\ieno, DRTL$_m$) on these backbones. As shown in Table. \ref{tab:SOTA}, for all five backbone models, the proposed DRTL-related schemes (DRTL$_p$ and DRTL$_m$) achieve the best performance on the real-world distorted test set, which indicates the effectiveness and generalization of our DRTL. Compared with the baseline scheme, the DRTL-enabled schemes could stably achieve nearly 0.3dB$\sim$0.5dB gains, which totally-fair results further reveal the effectiveness of our DRTL for few-shot image super-resolution. Moreover, for the simple-structured DnCNN, we find that it hardly converges well on the few-shot training clean-distorted real image pairs, and only achieves 29.297dB. In contrast, when applying our DRTL to DnCNN, this scheme could achieve large performance improvement (31.1dB in PSNR).

% Unlike previous works, which synthetic auxiliary distortion that is homologous to target distortion, our synthetic auxiliary distortion is heterogeneous to target real distortion, which cannot be syntheticed. 

Moreover, as shown in Table~\ref{tab:SOTA}, we also observe that with our selected seven distortions based on our distortion relation, the Pre-training and MAML schemes both achieve obvious gains in comparison with the baseline. But, they both ignore exploring how to better utilize the distortion relation for knowledge transfer. Thanks to a simple gradient modulation with distortion relation guidance, our DRTL-enabled schemes could achieve extra obvious gains compared with the Pre-training/MAML schemes when using different backbones.

In terms of subjective comparison, as shown in Figure~\ref{fig:subjective}, our method has the capability to restore more texture details for real-world distorted image, \egno, the buildings in the first row and the windows in the second row. We analyze that because our DRTL optimization strategy could transfer more valuable knowledge with distortion relation guidance from the auxiliary distortions to the target RealSR.
\begin{table}[htp]
\centering
\setlength{\tabcolsep}{4pt}{\begin{tabular}{c|c|c|c|c}
\hline%\toprule
Methods & DIP~\cite{ulyanov2018deepDIP} & Meta-ZSSR~\cite{soh2020metaZSSR} & Real-ESRGAN~\cite{wang2021realESRGAN} &   DRTL \\ \hline%\midrule
PSNR      &   30.32 & 29.68 & 28.25 &  \textbf{31.81}    \\
SSIM      &    0.870 & 0.856 & 0.860  &    \textbf{0.899}   \\
LPIPS     &    0.197 & 0.168 & \textbf{0.141} &    0.143   \\
\hline%\bottomrule
\end{tabular}}
\caption{Quantitative comparison of our DRTL with unsupervised, zero-shot, and  distortion-augmented methods.}

\label{tab:dip}
\end{table}

\subsubsection{Comparison with other Methods} we also select unsupervised method DIP~\cite{ulyanov2018deepDIP}, and zero-shot method Meta-ZSSR~\cite{soh2020metaZSSR} for comparison. For the distortion-augmented method, we compare our method with the famous Real-ESRGAN~\cite{wang2021realESRGAN}. 
 As shown in Table~\ref{tab:dip}, we can observe that DIP cannot work very well when processing real-world distortions. Real-ESRGAN~\cite{wang2021realESRGAN} achieves the best performance in LPIPS, while obtaining the worst performance in terms of PSNR and SSIM since the fake texture generation. Our DRTL on Few-shot RealSR can effectively avoid this problem.

%横向+纵向分析。  
% no pre-training (baseline), Pre-training, Meta-learning, Deep image prior, Learning invariant
%------results:
%%%%%%%% || directly training  || pre-training || Meta-learning || Ours || DIP || Learning invriant

\begin{figure}
	\centering
	\includegraphics[width=0.9\linewidth]{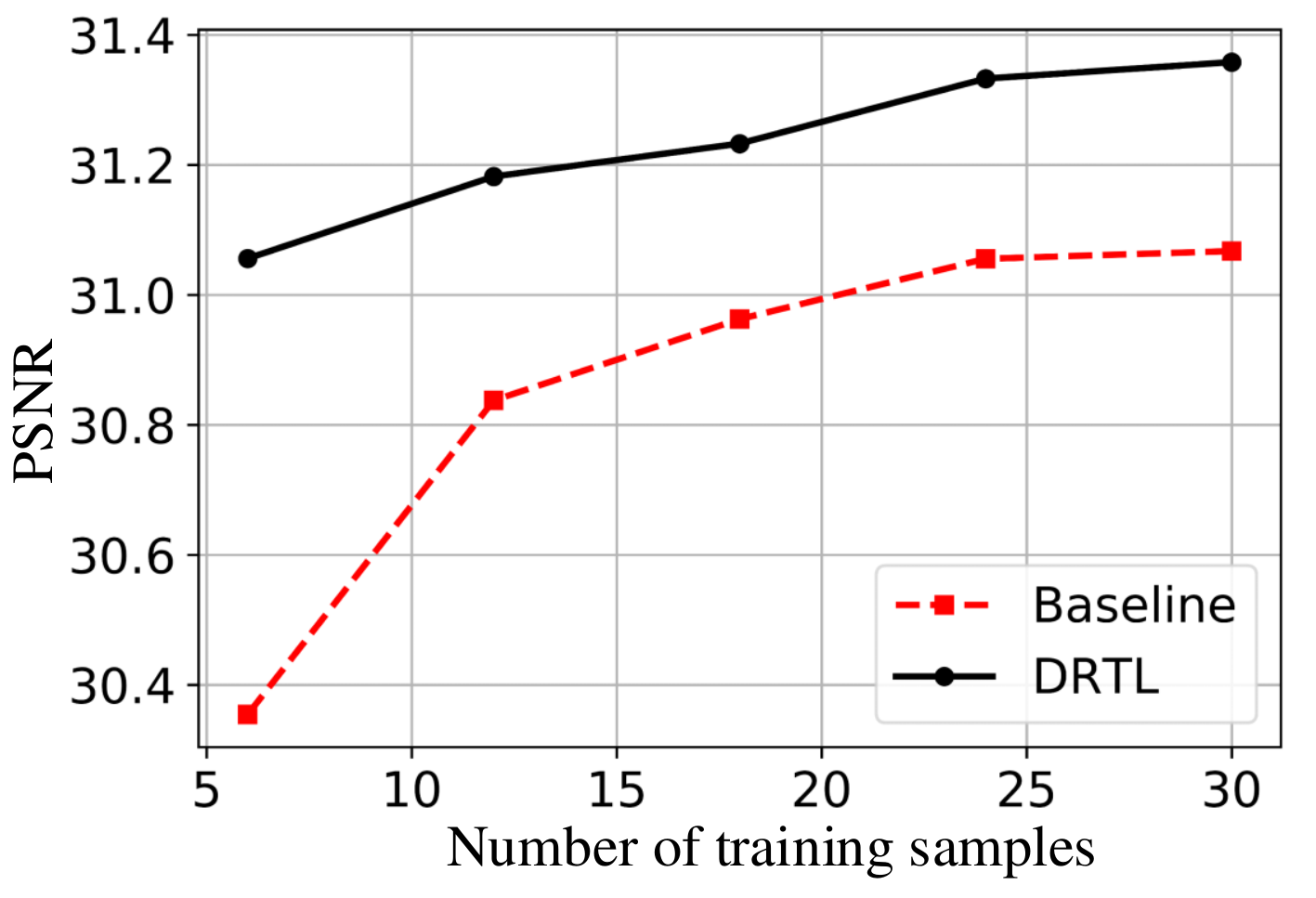}
	\caption{Influence of the number of real training data.}
	\label{fig:number}
\end{figure}

% 14关系图。-》随机
% graph similarity <-> relation (prior transfer衡量，即一类失真训练好的模型在另外失真上的性能来一定程度衡量similarity)

\subsection{Ablation Study}~\label{sec:ablation}

\subsubsection{Study on the Knowledge Memory Bank and Similarity Metric Choice.}
In our DRTL, we introduce a knowledge memory bank to better store the distortion priors. To validate its effectiveness and necessity, we remove the design from our distortion relation network and evaluate the performance of this scheme on the MAML framework. As shown in Table~\ref{tab:ab2}, the performance of scheme \emph{w/o Memory Bank} degrades since the correlation between different distortions cannot be captured effectively/accurately.
We also attempt to replace the cosine similarity with L1 or L2 distance to compute the edges of the distortion relation graph. However, as shown in Figure \ref{fig:Relation}, the relation coefficients do not satisfy a well-linear trend as cosine similarity did.

\subsubsection{Influence of the Number of Real Training Data.}
To study the influence of the number of real distorted/clean image training pairs w.r.t the final performance, we set several cases with the number of few-shot real image pairs respectively as 5, 10, 15, 20, 25, and 30 for comparison. As shown in Figure~\ref{fig:number}, with the number of samples going down, the performance of \emph{Baseline} scheme degrades quickly. In contrast, our DRTL scheme just decreases slightly. Moreover, our DRTL could achieve more gains when there are fewer samples, which further reveals the superiority of our DRTL method under the more challenging few-shot settings.

\begin{table}[htp]
\centering
\setlength{\tabcolsep}{14pt}{
\begin{tabular}{c|c|c|c}
\hline%\toprule
Methods         & PSNR & SSIM & LPIPS \\ \hline%\midrule
w Memory Bank   &   31.3670   &     0.8907  &   0.1476    \\
w/o Memory Bank &    31.2396  &  0.8879     &  0.1511     \\ 
\hline%\bottomrule
\end{tabular}
\caption{Quantitative comparisons of our proposed DRTL with/without memory bank, which is based on DRTL$_m$.}
\label{tab:ab2}}

\end{table}

\section{Conclusions}
\label{sec:conclusion}
In this paper, we are the first to have a close look at the challenging few-shot RealSR problem. Since the real clean-distorted image pairs are difficult to collect, we propose to transfer the task-relevant distortion knowledge from auxiliary synthetic distortions to real-world SR. However, the synthetic distortions and real distortion exist a large gap and the na\"ive transfer learning cannot be adaptively optimized with the distortion relations. Therefore, we propose the distortion relation graph with a prior knowledge memory bank to model the dependencies of different synthetic distortions and real-world RealSR. Based on the distortion relation graph, we could select the most relevant auxiliary distortions for the target RealSR. Moreover, we propose a distortion relation guided transfer learning (DRTL) framework with gradient reweighting for few-shot RealSR.  Extensive experiments on the few-shot RealSR have validated the effectiveness of our DRTL.

\bibliographystyle{IEEEtran}
\bibliography{references}

\end{document}